\pdfoutput=1
\documentclass{IEEE_lsens}
\usepackage[T1]{fontenc}
\usepackage{textcomp,amsmath,graphicx,array,booktabs,url}
\usepackage[cmintegrals]{newtxmath}
\usepackage{bm}
\usepackage[caption=false]{subfig}
\usepackage[noadjust]{cite}
\usepackage[hidelinks]{hyperref}
\hypersetup{pdftitle={Multimodal Sensor-Gated Mid-Air Launch and Vision-Based Recovery of a Child UAV on a Hovering Mothership Drone},pdfauthor={Aashish Sahu and R. Prasanth Kumar}}
\begin{document}
\markboth{IEEE Sensors Letters --- Manuscript for Review}{\thepage}
\IEEELSENSarticlesubject{Sensor Applications}
\title{Multimodal Sensor-Gated Mid-Air Launch and Vision-Based Recovery of a Child UAV on a Hovering Mothership Drone}
\author{
\IEEEauthorblockN{
Aashish~Sahu\IEEEauthorrefmark{1} and
R.~Prasanth~Kumar\IEEEauthorrefmark{2}
}
\IEEEauthorblockA{
\IEEEauthorrefmark{1}Department of Mechanical and Aerospace Engineering,
Indian Institute of Technology Hyderabad,\\
Kandi, Sangareddy 502284, India\\[0.4\baselineskip]
\IEEEauthorrefmark{2}Department of Artificial Intelligence,
Indian Institute of Technology Hyderabad,\\
Kandi, Sangareddy 502284, India
}
}

\IEEEtitleabstractindextext{

\begin{abstract}
Launching a micro unmanned aerial vehicle (UAV) from a hovering carrier and recovering it in flight requires reliable confirmation that the child is securely seated before the carrier is allowed to descend. This is challenging because vision provides relative guidance but does not directly indicate load transfer, and a downward-facing camera can lose its fiducial near touchdown. This letter presents a mothership--child quadrotor system in which mission transitions are gated by complementary nonvisual sensing. An upward-facing single-beam LiDAR on the recovery deck provides a binary occupancy state using hysteresis and a debounce interval derived from the child's occluding footprint and crossing speed; real-time kinematic (RTK) positioning conditions the carrier hover; and carrier thrust provides a proprioceptive indication of load transfer. In ten outdoor missions using an F450-class mothership and a Q250-class child with camera-guided recovery, nine completed successfully while RTK-fixed positioning was maintained through release and capture. A complementary MuJoCo study of fully autonomous fiducial-based recovery achieved touchdown 0.077~m from the deck center, while marker detection was lost within the final ${\approx}0.2$~m above the docked configuration. In simulation, carrier thrust changed by $-17.0$~N at release and $+17.3$~N at capture, agreeing within 6\% with the child's 18.0~N hover thrust. The results support a sensing architecture in which vision guides the approach, while range and proprioceptive information corroborate the final docking decision.
\end{abstract}

}
\maketitle
\section{Introduction}
Small multirotor UAVs are constrained by endurance and by repeated ground
takeoff and landing. A heterogeneous aerial system, in which a larger
mothership transports, launches, and recovers a smaller child UAV in
flight, can extend the child's operational range without requiring a ground
recovery site. Vision-based landing is widely used for short-range recovery
because it provides relative position without external infrastructure
\cite{xin2022review}. Prior work has demonstrated landing on moving
platforms using onboard perception \cite{falanga2017movingplatform},
model-predictive guidance for disturbance rejection \cite{feng2018mpc},
and image-based visual servoing for touchdown regulation
\cite{serra2016dynamic_ibvs}. Fiducials such as AprilTag
\cite{olson2011apriltag} and ArUco \cite{garrido2014aruco} simplify
terminal relative localization \cite{morales2023followland}. Related work from our group addressed system-level mother--child--UGV
coordination and aerial recovery \cite{sahu2026integrated}, and
RTK--vision/PPO-based autonomous recovery on an airborne carrier
\cite{sahu2026rtkppo}. In contrast, this letter isolates capture
confirmation as a sensing problem and formalizes LiDAR occupancy gating,
RTK-conditioned supervision, and thrust-based load-transfer corroboration.

An airborne carrier introduces a different sensing problem. First, the landing surface itself is moving and is disturbed by both child departure and recovery, so the relative motion depends on two aerial vehicles \cite{wang2022dualuav}. Second, the safety-critical variable is not only proximity to the deck but whether the child is \emph{docked}, defined here as seated on the deck and supported by the mothership. This state authorizes joint descent. Vision alone cannot directly verify it: fiducial detection indicates relative pose rather than mechanical support, the marker may leave the camera field of view near touchdown, and near-deck passes or partial contact can occur before secure capture.

This letter therefore formulates aerial recovery as a multimodal sensing problem and makes three contributions. 1)~An upward-facing single-beam LiDAR deck-occupancy detector using hysteresis and temporal debouncing, with the debounce interval selected from the child's occluding footprint and minimum crossing speed. 2)~A thrust-based proprioceptive capture cue that detects load transfer independently of the LiDAR and can distinguish a seated child from one hovering immediately above the deck. 3)~A quantitative analysis of the fiducial blind zone during autonomous recovery, together with outdoor and simulated mission results showing how nonvisual sensing complements vision at safety-critical release and capture events.

\section{System and Sensing Architecture}
\label{sec:system}

\begin{figure}[!t]
    \centering
    \includegraphics[width=\linewidth]{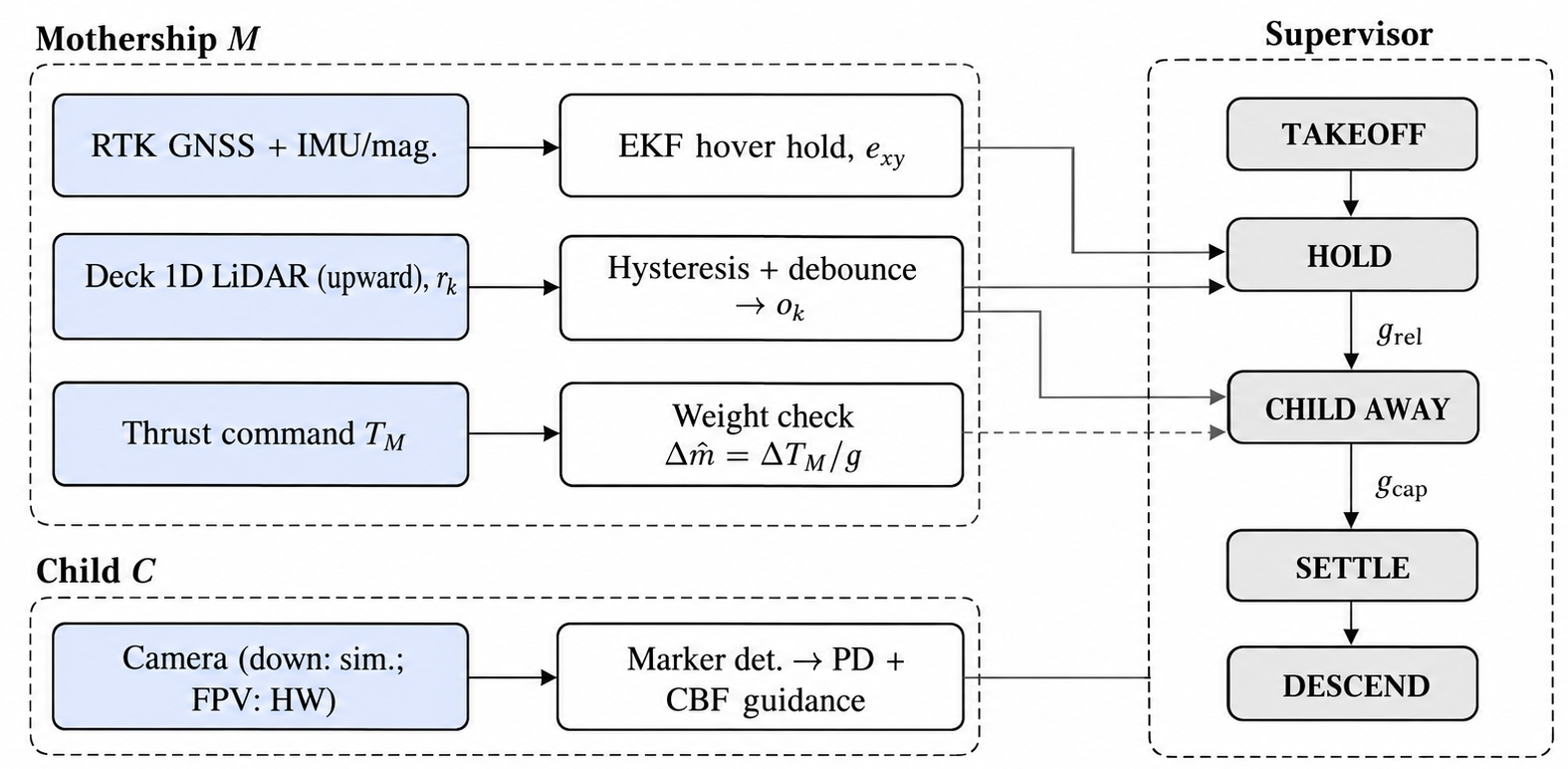}
    \caption{Sensing architecture and supervisor. Release guard $g_\mathrm{rel}$ requires RTK-fixed positioning, bounded hover error, and $o_k\!:\!1\!\to\!0$. Capture guard $g_\mathrm{cap}$ requires $o_k\!=\!1$ for $\tau_d$; thrust provides an independent load-transfer cue.}
    \label{fig:arch}
\end{figure}

\subsection{Platforms}
\begin{figure}[!t]
\centering
\subfloat[]{\includegraphics[height=2.15cm]{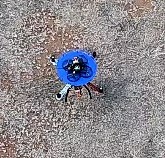}}\hfil
\subfloat[]{\includegraphics[height=2.15cm]{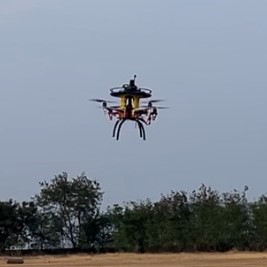}}\hfil
\subfloat[]{\includegraphics[height=2.15cm]{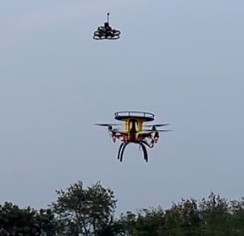}}\\[-1pt]
\subfloat[]{\includegraphics[height=2.15cm]{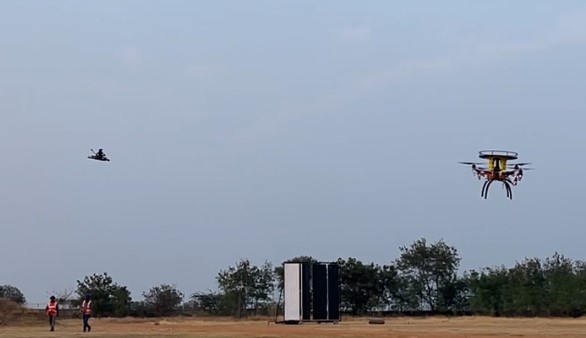}}\hfil
\subfloat[]{\includegraphics[height=2.15cm]{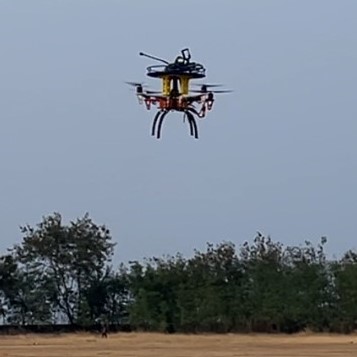}}\hfil
\subfloat[]{\includegraphics[height=2.15cm]{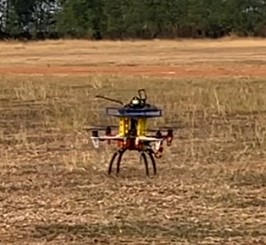}}
\caption{Outdoor mission. (a) Child seated on the instrumented deck. 
(b) Joint hover at 5~m. (c) Mid-air release. (d) Outbound flight while
the mothership holds position. (e) Camera-guided recovery with LiDAR
occupancy confirmation. (f) Autonomous joint landing.}
\label{fig:hw_seq}
\end{figure}

The mothership is an F450-class quadrotor (880-KV motors, $10\times4.5$-in propellers, 6800-mAh battery, mass $\approx1.5$~kg) running PX4 \cite{meier2015px4} with RTK GNSS \cite{sridhar2025swarm}. A 30-cm-diameter recovery deck is mounted 10~cm above the frame. The child is a Q250-class quadrotor ($\approx0.6$~kg) with an FPV camera. During outdoor tests, the mothership autonomously performs takeoff, 5-m hold, descent, and landing; the child is piloted from its live video, with the operator regulating deck image offset while the autopilot stabilizes attitude and thrust. Fig.~\ref{fig:hw_seq} shows one mission.

\subsection{Sensor Suite and Gating}
Fig.~\ref{fig:arch} summarizes the supervisor. RTK GNSS, inertial sensing, and magnetometer measurements condition mothership hover. Deck occupancy is measured by an upward-facing Benewake TFmini Plus single-point ToF LiDAR mounted near the deck center \cite{sahu2024hexacopter}. A docked child produces a persistent near-deck return, whereas a clear deck gives a distant or out-of-range measurement; the LiDAR is therefore used as a binary occupancy sensor.

While seated, $o_k=1$. After liftoff, $o_k\to0$ and the mothership holds its RTK position as the recovery target. The HOLD$\rightarrow$CHILD AWAY transition additionally requires RTK-fixed positioning and bounded hover error. Capture is declared only when $o_k=1$ persists for $\tau_d$, after which the mothership settles and descends. Dropouts or out-of-range returns are treated as empty, so descent remains inhibited without positive occupancy evidence.

\section{Sensing Methods}
\label{sec:methods}

\subsection{LiDAR Deck-Occupancy Detector}
Let $r_k$ be the LiDAR range and $o_k\in\{0,1\}$ the occupancy state. With $r_\mathrm{on}<r_\mathrm{off}$ and $N_d=\lceil\tau_d f_s\rceil$,
\begin{equation}
o_k=\begin{cases}
1, & o_{k-1}=0,\ r_i<r_\mathrm{on}\ \forall i\in\mathcal{N}_k,\\
0, & o_{k-1}=1,\ r_i>r_\mathrm{off}\ \forall i\in\mathcal{N}_k,\\
o_{k-1}, & \text{otherwise},
\end{cases}
\label{eq:occ}
\end{equation}
where $\mathcal{N}_k=\{k-N_d+1,\dots,k\}$. Hysteresis suppresses threshold chatter and debouncing rejects short occlusions.

Let $D_C$ be the longest child-body chord intersecting the beam. A fly-over at speed $v$ occludes it for at most $D_C/v$; hence
\begin{equation}
\tau_d>D_C/v_\mathrm{min}
\label{eq:debounce}
\end{equation}
rejects crossings with $v\ge v_\mathrm{min}$.

For the Q250, $D_C\approx0.25$~m and $v_\mathrm{min}=0.5$~m/s, giving $\tau_d>0.5$~s. We use $\tau_d=0.6$~s, $r_\mathrm{on}=0.10$~m, and $r_\mathrm{off}=0.20$~m. Because $r_\mathrm{on}$ lies near the TFmini Plus near-field region, the measurement is interpreted as persistent occupancy rather than precise contact range. A slow hover within $r_\mathrm{on}$ can still satisfy \eqref{eq:occ}, motivating an independent load-transfer cue.

\subsection{Proprioceptive Capture Verification}
Near hover,
$T_M\approx(m_M+o\,m_C)g$, so the thrust step estimates transferred mass as
\begin{equation}
\Delta\hat m=\Delta T_M/g.
\label{eq:dm}
\end{equation}
A hovering child gives $\Delta\hat m\approx0$, whereas a seated child gives $\Delta\hat m\approx m_C$, resolving the main ambiguity of occupancy sensing. On hardware, however, $T_M$ is only a normalized PX4 command and requires motor--propeller and voltage-dependent calibration; it is therefore used qualitatively, while quantitative verification of \eqref{eq:dm} is performed in simulation.

\subsection{Vision-Based Relative Localization and Blind Zone}
In the autonomous configuration, the child detects a deck fiducial using a downward camera \cite{olson2011apriltag,garrido2014aruco}. The planar error is $e_{xy}=\|\mathbf{p}^{xy}_C-\mathbf{p}^{xy}_D\|$. A square marker of side $s$ is fully visible only when
\begin{equation}
h_\mathrm{min}=\frac{s}{2\tan(\theta/2)},
\label{eq:blind}
\end{equation}
where $\theta$ is camera field of view. For $s=0.20$~m and $\theta=65^\circ$, $h_\mathrm{min}\approx0.16$~m. Detection therefore degrades near touchdown as the marker fills the image. Multi-scale markers \cite{romero2019fractal} can reduce this blind region, but vision still measures relative pose rather than mechanical support; final docking is therefore assigned to the nonvisual sensing channels.
\section{Experimental Validation}
\label{sec:results}

\subsection{Outdoor Flights}
Ten end-to-end missions comprising joint takeoff, 5-m hover, child release,
outbound flight, recovery, LiDAR-confirmed occupancy, and joint landing were
conducted; nine completed successfully. The unsuccessful trial coincided
with gust disturbance and noisy sensing. In every successful mission,
$o_k$ transitioned once following release and once following recovery, and
no descent was initiated before the occupancy condition persisted for the
required debounce interval. With the deck clear, the upward-facing TFmini
Plus returned a distant or out-of-range state; recovery produced a
persistent near-deck return. Because the docked geometry lies close to the
sensor's near-field region, the LiDAR is treated as a binary occupancy
sensor rather than as a precision contact-range measurement.

The GNSS log showed 27 satellites in use, with horizontal and vertical
dilution of precision below 1. The RTK-fixed solution was maintained
through takeoff, release, and recovery. Brief degradations to RTK float and DGPS occurred only after capture and did not disturb joint descent. Thus, RTK-fixed positioning was
available at both safety-critical events, and the RTK condition of
$g_\mathrm{rel}$ was satisfied at child release.

\begin{figure}[!t]
\centering
\includegraphics[width=0.96\columnwidth]{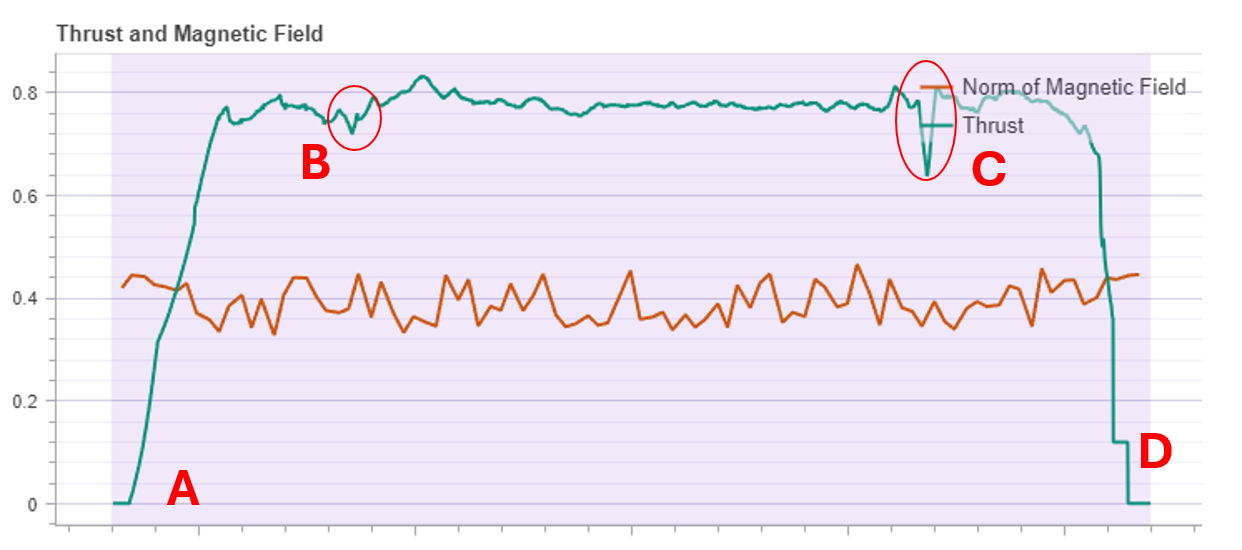}
\caption{Mothership normalized thrust command and magnetic-field magnitude
during a representative outdoor mission. A: takeoff; B: child release;
C: recovery/landing disturbance; D: joint touchdown.}
\label{fig:thrust}
\end{figure}

In Fig.~\ref{fig:thrust}, the normalized thrust command is approximately
0.40 during joint hover. Child release (B) produces a transient decrease
to about 0.35, while the recovery/landing interval (C) contains a deeper
transient to about 0.32 before settling. These features provide
proprioceptive event cues independent of the LiDAR, but are not interpreted
as calibrated force measurements. For the hardware masses,
\eqref{eq:dm} predicts a load transfer of $m_Cg\approx5.9$~N, approximately
29\% of the joint-hover weight. Conversion of normalized PX4 thrust to
$\Delta\hat m$ requires motor--propeller and voltage-dependent calibration;
therefore, the hardware thrust signal is used here only qualitatively.
The magnetic-field magnitude remained within 0.42--0.47~G without
event-correlated excursions.

\subsection{Autonomous Vision-Based Recovery in MuJoCo}
To evaluate autonomous recovery, the mission was reproduced in MuJoCo
\cite{todorov2012mujoco} with PX4-style telemetry. The mothership holds at
10~m while the child uses a downward-facing camera to detect the deck
fiducial and track it using PD guidance protected by a control-barrier-
function safety filter \cite{ames2019cbf}. The safety filter remained
inactive in the reported run because the nominal trajectory satisfied its
constraint. The simulated vehicles are not scaled to the hardware:
mothership hover thrust is 137.3~N and child hover thrust is 18.0~N.
Mission phases are carried takeoff (TO), release (RL), survey (SV), return
(RT), terminal approach (TM), settle (ST), and joint descent (DS).

\begin{figure}[!t]
\centering
\includegraphics[width=\columnwidth]{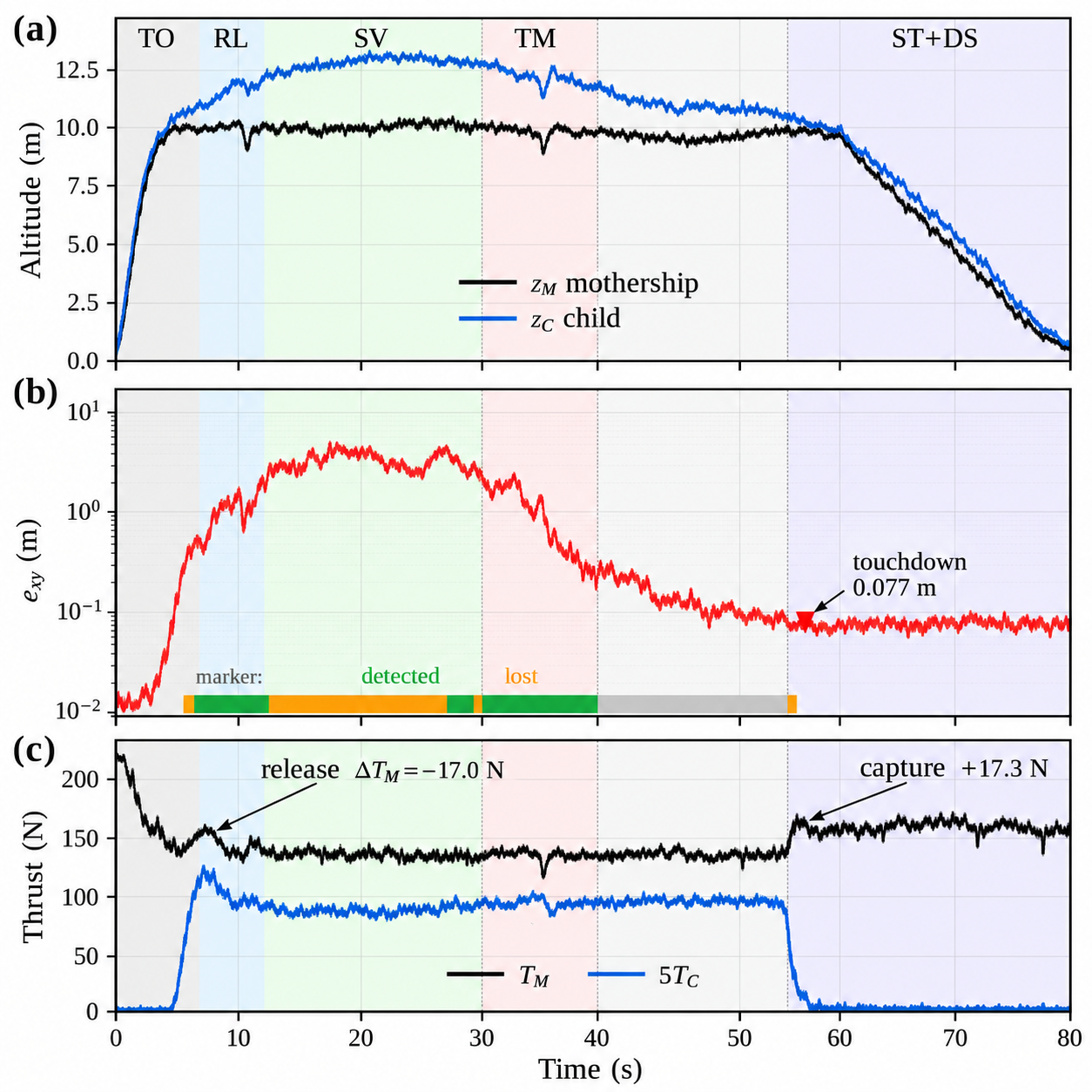}
\caption{MuJoCo telemetry. (a) Mothership and child altitudes.
(b) Planar error $e_{xy}$ and marker-detection state.
(c) Mothership thrust $T_M$ and scaled child thrust $T_C$.
Background colors indicate mission phases.}
\label{fig:simtel}
\end{figure}

\begin{figure}[!t]
\centering
\subfloat[$h\!\approx\!0.9$~m, detected]
{\includegraphics[width=0.48\columnwidth]{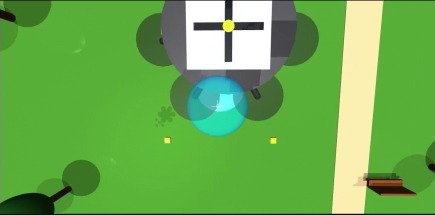}}\hfil
\subfloat[$h\!\approx\!2.1$~m, detected]
{\includegraphics[width=0.48\columnwidth]{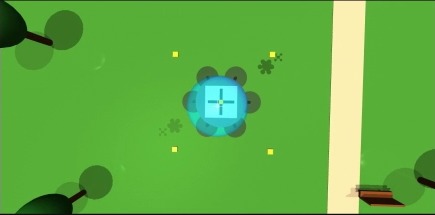}}\\[-1pt]
\subfloat[$h\!\approx\!0.25$~m, detected]
{\includegraphics[width=0.48\columnwidth]{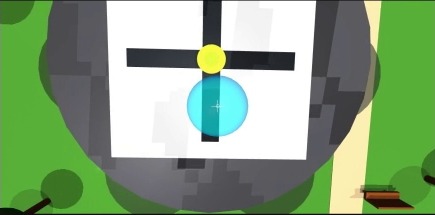}}\hfil
\subfloat[$h\!\approx\!0.09$~m, lost]
{\includegraphics[width=0.48\columnwidth]{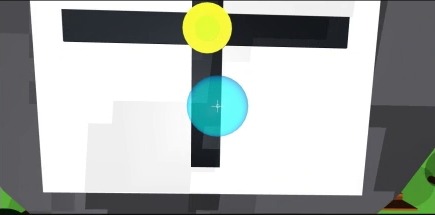}}
\caption{Child downward-camera view in MuJoCo; $h$ denotes height above
the docked configuration. (a) Reacquisition after release.
(b) Terminal-approach initialization. (c) Last detection near touchdown.
(d) Marker overfills the field of view and detection is lost.}
\label{fig:cam}
\end{figure}

\begin{table}[!t]
\caption{Performance Summary}
\label{tab:metrics}
\centering
\renewcommand{\arraystretch}{1.08}
\begin{tabular*}{\columnwidth}{@{}l@{\extracolsep{\fill}}r@{}}
\hline
Metric & Value \\
\hline
\multicolumn{2}{@{}l}{\emph{Outdoor flights (10 missions)}}\\
End-to-end mission success & 9/10 \\
Deck occupancy sensor & TFmini Plus, $\tau_d=0.6$ s \\
GNSS at release and recovery & RTK fixed, 27 sat. \\
\multicolumn{2}{@{}l}{\emph{MuJoCo autonomous recovery (single run)}}\\
Release / dock / mission end & 6.5 / 55.2 / 78.3 s \\
Mothership hover planar RMS error & 0.04 m \\
Terminal planar RMS error & 0.15 m \\
Planar error at touchdown & 0.077 m \\
Go-arounds & 3 \\
Marker blind zone & $\approx0.2$ m \\
Thrust step, release / capture & $-17.0$ / $+17.3$ N \\
Altitude sag at capture & 0.11 m \\
Peak tilt near docking & 3.2$^\circ$ \\
\hline
\end{tabular*}
\end{table}

Fig.~\ref{fig:simtel} and Table~\ref{tab:metrics} summarize the autonomous
run. The mothership reached hold at 4.5~s and released the child at 6.5~s.
During post-release climb, marker detection was regained 0.33~m above the
deck. It was lost during the outbound survey ($e_{xy}$ up to 5.6~m) and
reacquired on return. Outside the go-arounds, terminal $e_{xy}$ remained
between 0.02 and 0.10~m; including the go-arounds gives an RMS error of
0.15~m. Detection was last available at $h\approx0.25$~m and was lost by
$h\approx0.09$~m [Fig.~\ref{fig:cam}(c),(d)]. During climb, the corresponding
transition occurred between 0.16 and 0.33~m. The observed blind region of
approximately 0.2~m is therefore consistent with the geometric prediction
of \eqref{eq:blind}. The child completed the final segment without visual
feedback and touched down 0.077~m from the deck center after three
go-arounds.

At release, $T_M$ decreased from 154.5 to 137.5~N
($\Delta T_M=-17.0$~N) and, following capture, settled at 154.6~N,
corresponding to a $+17.3$-N load increase. The child's hover thrust during
return was $18.0\pm0.3$~N; thus the carrier thrust step agrees with the
child weight within 6\%. Although this agreement follows naturally from
simulation force balance, it demonstrates that the load-transfer event is
observable independently of vision. The step was resolved within 2.5~s of
capture; during this interval the 0.11-m altitude sag recovered to within
0.02~m before joint descent. Peak mothership tilt near docking was
3.2$^\circ$.

\subsection{Sensor-to-Action Mapping}
Table~\ref{tab:events} summarizes the complementary sensor signatures.
In the present hardware implementation, RTK and LiDAR provide the actual
supervisor gates, while thrust is used only as a corroborative cue and is
quantitatively evaluated in simulation. Vision becomes unavailable in the
near-deck blind region; the LiDAR cannot measure lateral displacement or
distinguish contact from a sufficiently slow low hover; and the thrust cue
becomes informative only after load transfer. Thus, the channels provide
complementary rather than redundant information. Under \eqref{eq:occ}, a
fast pass shorter than $\tau_d$ leaves $o_k=0$, while a slow low hover is
the principal ambiguity requiring load-transfer corroboration.

\begin{table}[!t]
\caption{Sensor Evidence and Mothership Action per Event}
\label{tab:events}
\centering
\renewcommand{\arraystretch}{1.08}
\setlength{\tabcolsep}{2.5pt}
\begin{tabular*}{\columnwidth}{@{}l@{\extracolsep{\fill}}cccl@{}}
\hline
Event & LiDAR $o_k$ & Marker & $\Delta\hat m$ & Action\\
\hline
Child seated & 1 & -- & -- & Joint takeoff/hover\\
Release & $1\!\to\!0$ & blind & $-m_C$ & Hold position\\
Child away & 0 & lost/seen & 0 & Descent inhibited\\
Fast deck pass & stays 0 & blind & 0 & Keep holding\\
Low hover & may rise & blind & $\approx0$ & Needs corroboration\\
Touchdown & $0\!\to\!1$ & blind & $+m_C$ & Settle, descend\\
\hline
\end{tabular*}
\end{table}

A single-beam ToF ranger is attractive for this role because it is
lightweight, contact-free, and produces one scalar measurement per sample,
allowing \eqref{eq:occ} to execute without image processing. Facing upward,
it observes the docking volume directly and is insensitive to illumination,
motion blur, and marker overfill. An off-center child may not continuously
occlude the beam, preventing premature descent. Contact switches or load
cells can also detect touchdown but must mechanically carry impact loads,
whereas the LiDAR remains non-contact.

\subsection{Limitations}
The hardware and simulation studies are complementary rather than
equivalent. Hardware experiments validate mid-air release, piloted
camera-guided recovery, RTK-conditioned hover, and LiDAR-based deck
occupancy on real vehicles. MuJoCo evaluates autonomous fiducial recovery
and quantitative thrust-based load-transfer sensing on a vehicle model not
scaled to the hardware. The autonomous simulation result is from a single
run, and the hardware thrust command has not yet been calibrated to force
or incorporated into the supervisor gate.

\section{Conclusion}
This letter presented a mothership--child UAV architecture in which
mid-air launch and recovery are supported by complementary nonvisual
sensing. An upward-facing TFmini Plus LiDAR with hysteresis and temporal
debouncing determines persistent deck occupancy, RTK GNSS conditions the
carrier hover, and carrier thrust provides a proprioceptive load-transfer
cue. Nine of ten outdoor missions completed successfully with RTK-fixed
positioning during release and recovery. In MuJoCo, autonomous
fiducial-based recovery achieved a 0.077-m touchdown error, exposed an
approximately 0.2-m near-deck visual blind zone, and produced carrier
thrust steps agreeing with the child weight within 6\%. The results support
using vision primarily for relative guidance while nonvisual sensing
provides independent evidence for the final docking decision. Future work
will calibrate the hardware thrust channel, integrate it with LiDAR
occupancy for probabilistic capture detection, move autonomous fiducial
guidance onto the physical child UAV, and evaluate repeated trials under
wind and degraded GNSS.

\end{document}